\documentclass[10pt,twocolumn,letterpaper]{article}

\usepackage[pagenumbers]{cvpr} % To force page numbers, e.g. for an arXiv version
\usepackage[accsupp]{axessibility} % Improves PDF readability for those with disabilities.

\usepackage{graphicx}
\usepackage{amsmath}
\usepackage{amssymb}
\usepackage{booktabs}
\usepackage[table,xcdraw, dvipsnames]{xcolor}
\usepackage{pifont}
\newcommand*\colourcheck[1]{%
  \expandafter\newcommand\csname #1check\endcsname{\textcolor{ForestGreen}{\ding{52}}}%
}
\colourcheck{green}

\usepackage{soul}

\usepackage{textcomp}

\newcommand{\cmark}{\ding{51}}

\usepackage{multirow}
\usepackage{newfloat}
\usepackage[normalem]{ulem}
\useunder{\uline}{\ul}{}

\usepackage{hyperref}
\hypersetup{colorlinks, urlcolor=magenta, citecolor=green}
\usepackage[dvipsnames]{xcolor}
\usepackage[capitalize]{cleveref}
\crefname{section}{Sec.}{Secs.}
\Crefname{section}{Section}{Sections}
\Crefname{table}{Table}{Tables}
\crefname{table}{Tab.}{Tabs.}

\def\cvprPaperID{2} % *** Enter the CVPR Paper ID here
\def\confName{CVPR}
\def\confYear{2023}

\begin{document}

\title{Best Practices for 2-Body Pose Forecasting}

% \and

% \author{Muhammad Rameez Ur Rahman$^*$\space\space\space Luca Scofano$^*$ \space\space\space Edoardo De Matteis \\ 
% Alessandro Flaborea \space\space\space Alessio Sampieri \space\space\space Fabio Galasso\\
% Sapienza University of Rome, Italy \space\space\space \\
% {\tt\small {rahman,scofano,dematteis,flaborea,sampieri,galasso}@di.uniroma1.it}}

\author{Muhammad Rameez Ur Rahman$^{*, 1}$ \and Luca Scofano$^{*,2}$ \and Edoardo De Matteis$^1$ \and Alessandro Flaborea$^1$ \and Alessio Sampieri$^2$ \and Fabio Galasso$^1$ \\
Sapienza University of Rome, Italy \\
{\tt\small $^1$\{rahman, dematteis, flaborea, galasso\}@di.uniroma1.it} \\
{\tt\small $^2$\{scofano, sampieri\}@diag.uniroma1.it}
% \and
}

\maketitle
\begin{NoHyper}
    \def\thefootnote{*}\footnotetext{Equal contribution.}
    \def\thefootnote{\arabic{footnote}}
\end{NoHyper}

\def\ckmk{\tikz\fill[scale=0.4](0,.35) -- (.25,0) -- (1,.7) -- (.25,.15) -- cycle;} 

\newcommand{\ls}[1]{\color{ForestGreen}#1}
\newcommand{\LS}[1]{{\color{ForestGreen}{\ls [Luca: #1]}}}

\newcommand{\ra}[1]{\color{BrickRed}#1}
\newcommand{\RA}[1]{{\color{BrickRed}{\ra [Rameez: #1]}}}

\newcommand{\ed}[1]{\color{Cerulean}#1}
\newcommand{\ED}[1]{{\color{Cerulean}{\ed [Edoardo: #1]}}}

\newcommand{\as}[1]{\color{blue}#1}
\newcommand{\AS}[1]{{\color{blue}{\as [Alessio: #1]}}}

\newcommand{\fg}[1]{\color{orange}#1}
\newcommand{\FG}[1]{{\color{orange}{\fg [Fabio: #1]}}}

\newcommand{\af}[1]{\color{purple}#1}
\newcommand{\AF}[1]{{\color{purple}{\af [AF: #1]}}}

\newcommand{\modelname}{IFS-GCN~}

\maketitle

% \LS{Comment from Luca}\\
% \RA{Comment from Rameez}\\
% \ED{Comment from Edoardo}\\
% \AS{Comment from Alessio}\\
% \FG{Comment from Fabio}\\
% \AF{Comment from Ale}\\

%%%%%%%%% ABSTRACT
\begin{abstract}

The task of collaborative human pose forecasting stands for predicting the future poses of multiple interacting people, given those in previous frames.
Predicting two people in interaction, instead of each separately, promises better performance, due to their body-body motion correlations. But the task has remained so far primarily unexplored.

In this paper, we review the progress in human pose forecasting and provide an in-depth assessment of the single-person practices that perform best for 2-body collaborative motion forecasting.
Our study confirms the positive impact of frequency input representations, space-time separable and fully-learnable interaction adjacencies for the encoding GCN and FC decoding.
Other single-person practices do not transfer to 2-body, so the proposed best ones do not include hierarchical body modeling or attention-based interaction encoding.

We further contribute a novel initialization procedure for the 2-body spatial interaction parameters of the encoder, which benefits performance and stability. 
Altogether, our proposed 2-body pose forecasting best practices yield a performance improvement of 21.9\% over the state-of-the-art on the most recent ExPI dataset, whereby the novel initialization accounts for 3.5\%.
See our project page at 
\href{https://www.pinlab.org/bestpractices2body}{https://www.pinlab.org/bestpractices2body}
\end{abstract}

%%%%%%%%% INTRODUCTION
\section{Introduction}
\label{sec:introduction}

Human 2-body pose forecasting predicts the future body poses of two people in interaction jointly. The task is relevant to long-term pose tracking \cite{Andriluka2018PoseTrackAB}, to understanding interacting pairs in sports such as dancing~\cite{guo21} and to the collaborative assembly in industry~\cite{Duarte18,koppula21}, towards human-robot collaboration~\cite{sampieri22}.
Considering the concurrent prediction of two bodies helps in cases where the people act synergistically.
However, this task has remained mostly unexplored and limited to the dataset of \cite{guo21}\footnote{Beyond \cite{guo21}, another multi-body dataset has been introduced by~\cite{Fieraru_2020_CVPR}, but annotations are only available for one individual at the time of writing.}.
Also, this differs from the related task of human trajectory forecasting, where  social interaction has been key to most recent progress~\cite{monti21dag, Kothari22, yuan2021agent, Abduallah22}.

\begin{figure*}[t]
    \centering
    \includegraphics[width=\textwidth]{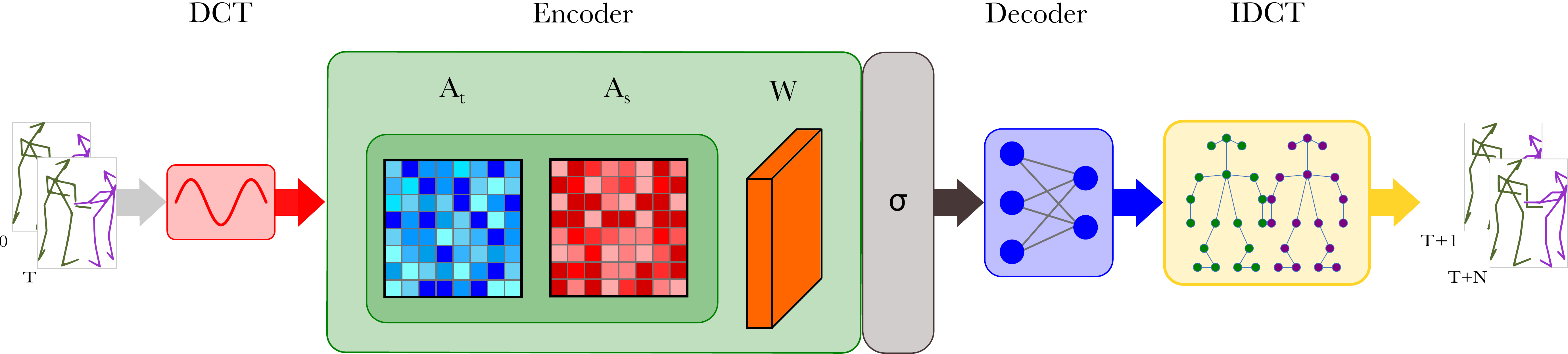}
    \caption{
    The general architecture of a 2-body pose forecasting model employing best practices.
    First, 3D joint coordinates are mapped to frequencies by DCT coefficients, a best input representation practice.
    Secondly, body kinematics are encoded by layers of a GCN $\sigma( A_s A_t X W)$, with separable space-time adjacency matrices $\sigma(A_t,A_s)$, learned unconstrainedly, upon our proposed parameter initialization.
    Thirdly, the FC-based decoder outputs future poses for the two people, mapped to 3D coordinates with inverse-DCT (IDCT).}
    \label{fig:teaser}
\end{figure*}

There has been vast progress in single human pose forecasting~\cite{Dang21, Ma22, guo2022back}, which has not transferred to the 2-body counterpart.
Single-person techniques~\cite{benzine19, Dabral19, corona21} tested on two-people data underperform, which is unsurprising, as they neglect the body-body motion correlations~\cite{guo21}.
This motivates the current work, where the most recent modeling advancements are analyzed and integrated.
Here, we refer to the best and complementary modeling aspects as \emph{best practices}, which we leverage to bootstrap research on 2-body forecasting.

We propose a systematic analysis of single-person skeleton-based best practices by considering three processing stages (cf. Fig.~\ref{fig:teaser}): input representation, encoding, and decoding.
For the first stage, we identify Discrete Cosine Transform (DCT)~\cite{cai20, guo21, mao20his, mao19ltd, mao21multi} as an asset to cope with the periodic body movements.
For the second stage, we set to encode the body kinematics by Graph Convolutional Networks (GCN), which power the vast majority of most recent techniques~\cite{mao19ltd, mao20his, mao21multi, guo21, sofianos21, guo2022back} and subsume general MLP-based formulations~\cite{guo2022back}.
Here we evaluate as best practices the separability of space and time dynamics~\cite{sofianos21}, the learnable adjacencies versus kinematic trees~\cite{yan18}, attention~\cite{guo21}, and hierarchical body representations~\cite{Dang21}. 
Finally, for the third stage, we contrast the widely-adopted~\cite{sofianos21, Ma22, sampieri22} decoding with convolutional networks (\textit{a.k.a.} Temporal Convolutional Network--TCN~\cite{bai2018empirical}) with the simpler Fully Connected (FC) layers~\cite{guo2022back}.

We propose a novel initialization technique for the learnable GCN parameters in the encoder. 
A large body of literature asserts the importance of initialization for performance, convergence speed, and robustness, and theory has been devised for MLP~\cite{glorot10} and ConvNets~\cite{he15, Krhenbhl2016}.
Up until recently, there has been a limited necessity for ad-hoc GCN initialization theories since techniques leveraged mainly shallow networks with fixed graphs structures (e.g., \ the people neighbors\cite{Atwood16, Li18, tong20}, the kinematic tree~\cite{yan18, Dang21}) or spectral normalizations~\cite{kipf17, kipf18nri}.
Since we determine that unconstrained learnable GCN affinities are best practices, we also develop a novel theory (See Sec.~\ref{ssec:init}) and experimental study (See Sec.~\ref{ssec:init_discussion}) on the initialization of GCN parameters.

Integrating the selected best practices into a 2-body pose forecasting model yields a large-margin improvement of 21.9\% \emph{wrt} the state-of-the-art (SoA) on the most recent ExPI dataset~\cite{guo21}.
The best-practice model is also 5 times faster than the current best technique and only has 2\% of its parameters (cf.\ Table~\ref{tab:interaction}).
The improvement is similarly consistent in generalization tests, across unseen actions with an overall improvement of 14.7\% (cf.\ Table~\ref{tab:unseen}) and 14.2\% for unseen actors (cf.\ Table~\ref{tab:single}). 
And the same best-practice model performs on par (cf.\ Table~\ref{tab:h36m}) with the leading single-person pose forecasting techniques on the established Human3.6M dataset~\cite{h36m}, without any hyper-parameter tuning.
The novel initialization, proposed for the unconstrained learning of GCN affinities, contributes an average performance improvement of 3.5\%, and it increases stability, as it reduces the long-term forecasting performance variance by (at least) a factor of 2.

% List of contributions
The main contributions are summarized as follows:
\begin{itemize}
    \item We thoroughly evaluate all leading best practices from single-person pose forecasting and bootstrap research on the 2-body task counterpart;
    \item We propose a novel theory and experimental study on the initialization of GCNs, applying to unconstrained learnable affinities, accounting for an increase in performance of 3.5\% and a 2-fold increase in stability;
    \item On a closed-set dataset configuration, the best-practice model outperforms the 2-body forecasting SoA by a large margin of 21.9\% while employing 2\% of the parameters and running 5 times faster.
\end{itemize}

%%%%%%%%% RELATED WORK
\section{Related Work}
\label{sec:related}

Here we review related work from the field of human pose forecasting, specifically approaches of spatio-temporal pose modeling and hierarchical body representations. Additionally, we review relevant literature from initialization and multi-agent trajectory forecasting.

\paragraph{Human pose forecasting.} Established methodologies for (single) human pose forecasting include Temporal Convolutional Network~\cite{Li18}, Recurrent Neural Network~\cite{mao19ltd, Fragkiadaki15, Wang19, Ma22} and Transformer Networks~\cite{guo21, Aksan21}. The MLP-based approach of \cite{guo2022back} holds SoA performance. 

Graph Neural Networks (GCN)~\cite{kipf17,yan18} are most popular on the task~\cite{Dang21,Li2022SymbioticGN,sampieri22}, due to their simplicity and effectiveness. GCNs model the kinematic body part interactions by a plain adjacency matrix at a fraction of the parameters of the otherwise required attention mechanism~\cite{guo21,mao20his}.
In this realm, \cite{mao20his} integrates DCT to consider motion frequency; \cite{Dang21,Li_2020_CVPR} adopt multi-scale hierarchical representations, grouping joints to model relations between coarser body parts; \cite{sofianos21,sampieri22} factorize the spatial and temporal adjacency matrices, and they propose to learn them, unconstrainedly, without kinematic tree priors nor spectral normalization.

As we know, the only work that addresses multi-body pose forecasting is \cite{wang21}. However, they utilize datasets that do not contain highly interactive actions. For comparison, we ran their model with our setup as a comparison with our proposed method (See Tab.~\ref{tab:expi}).
By contrast, for the task of 2-body pose forecasting, \cite{guo21} provides the solely-available dataset (ExPI) and the only 2-body-specific technique, adaptation of \cite{mao20his} with cross-person attention. Not surprisingly, this outperforms single-person techniques. 

\paragraph{Initialization.}
A proper initialization improves performance and accelerates convergence~\cite{Krizhevsky12}, limiting vanishing and exploding gradients~\cite{glorot10,he15}.
Techniques have been concerned with initializing the weights of linear~\cite{glorot10} and convolutional~\cite{he15, Krhenbhl2016, Mishkin15} layers, generalizing from hyperbolic (tanh) to rectified-linear unit (ReLU) activations.
For GCNs, spectral techniques~\cite{kipf17,zhang2021magnet,li_2018_agcn} rely on the spectral normalization of the adjacency matrix to elude vanishing and exploding gradients, while spatial techniques~\cite{Atwood16} resort to degree-normalized transition matrices, derived from the adjacency.
In all prior study cases, the graph connectivity is given.
To the best of our knowledge, this work presents the first theoretical and empirical analysis of GCN initialization in the case of unconstrained learnable graph connectivity and edge weights.

\paragraph{Multi-agent trajectory forecasting.}
For trajectory forecasting, employed techniques include attention~\cite{yuan2021agent, Kothari22, Huang19} and graph-based modeling~\cite{Li2020EvolveGraphMT, Shi21, monti21dag}. The multi-agent relations may parallel the joint-joint interaction. However, nodes in a graph of joints have a fixed cardinality and a semantic meaning (head, torso, hand, etc.), which does not apply to general agent-agent graphs. 
Notably, best trajectory forecasting techniques model the agent-agent interaction~\cite{Li2020EvolveGraphMT,yuan2021agent,Shi21,Kothari22,Huang19,monti21dag}, which aligns with the motivation of this work, to forecast the poses of people jointly.

%%%%%%%%% METHODOLOGY
\section{Methodology}\label{sec:methodology}

We explore the best models for single-body pose forecasting~\cite{mao19ltd, mao20his, sofianos21, Dang21, guo2022back} and select best practices for the 2-body task.
We group and evaluate practices in three processing stages (cf.\ Fig.~\ref{fig:teaser}): 1) input representation (Sec.~\ref{ssec:inp_rep}); 2) encoding of the body kinematics in the observed frames (Sec.~\ref{ssec:enc}); 3) decoding of the future poses (Sec.~\ref{ssec:dec}). In Sec.~\ref{ssec:init}, we provide a theory for the proposed unconstrained-GCN initialization. To facilitate reading, we mark with a green check \greencheck\ the selected \emph{best practices} upon evaluation, cf.\ Sec.~\ref{sec:results}.

\paragraph{Problem formalization.} \label{par:prob}
Across $T$ frames, we observe the motion of two human bodies $\mathcal{B}^1$ and $\mathcal{B}^2$, each consisting of $J$ three-dimensional joints.
At time $t$, the 3D body pose of each person is given by corresponding tensors $\mathcal{B}_{t}^{1}, \mathcal{B}_{t}^{2} \in \mathbb{R}^{3 \times J}$. 
We define the concatenation of two bodies at timeframe $t$ as $\textbf{x}_t = \mathcal{B}_t^1 || \mathcal{B}_t^2$, thus the observed motion history in $T$ frames is $\mathcal{X}_{in} = \left[ \textbf{x}_1, \dots, \textbf{x}_T \right] \in \mathbb{R}^{T \times 3 \times 2J}$. 
Our goal is to predict the future $N$ frames' poses $\mathcal{X}_{out} = \left[ \textbf{x}_{T+1}, \dots, \textbf{x}_{T+N} \right] \in \mathbb{R}^{N \times 3 \times 2J}$.

\paragraph{Preliminaries on the encoder-decoder baseline.} 
We adopt an encoder-decoder architecture~\cite{sofianos21, sampieri22}, and following \cite{yan18, zhang17}, we encode the observed body parts and their kinematic interaction through a GCN, defined as 
\begin{equation}
    \label{eq:plaingcn}
    {\mathbf{Y}} = \sigma({A}\mathbf{X}{W}),
\end{equation}
where $A$ is the adjacency matrix, $W$ learnable weights and $\sigma$ an activation function.
Other encodings such as RNNs~\cite{yong15, chiu19} and MLPs~\cite{guo2022back} have been proposed, whereas we opt for a graph-based model to exploit the non-euclidean nature of graphs.
As a decoder, we examine either a single fully connected layer as in \cite{guo2022back} or a convolutional architecture~\cite{sofianos21, Ma22}.

\subsection{Input Representation}\label{ssec:inp_rep}

Most recent techniques~\cite{guo2022back, mao19ltd, mao20his, Akhter08} use Discrete Cosine Transform (DCT) to represent 3D coordinate input as frequencies, under the claim that this captures the dynamic patterns of moving people better.

\paragraph{Frequency encoding} \label{para:dct} \greencheck \\
Given the $j$-th body joint and the $t$-th timeframe we define the $i$-th DCT coefficient as 
\begin{gather}
    \mathcal{F}(\mathcal{X}^{in})_{j,i} = \sqrt{\frac{2}{T}} \sum_{t=1}^T x_{j,t}  \frac{1}{\sqrt{1+\delta_{i1}}} \cos \left(\alpha\right) \\
    \alpha = \frac{\pi}{2T}(2t-1)(i-1),
\end{gather}
% \AF{with $x_{j,t}$ as ...} and $\delta$ being the Kronecker delta function:
% \begin{equation}
%     \delta_{ij}=
%     \begin{cases}
%             1 &         \text{if } i=j,\\
%             0 &         \text{if } i\neq j.
%     \end{cases}
% \end{equation}

where the Kronecker delta function $\delta_{ij} \in {0,1}$  has null value if $i \neq j$ and 1 otherwise.
After inference, frequencies are remapped to the pose representation via the inverse DCT decoding function $\mathcal{F}^{-1}$.
Previous works~\cite{mao20his, mao21multi} truncate high frequencies to avoid jittery motion; we consider the impact of the number of retained DCT coefficients and discover that employing all of them yields the best performances.
Studies on the impact of DCT coefficients are shown in Sec.~\ref{sec:results} and Table~\ref{tab:interaction}.

\subsection{Encoding Best Practices}\label{ssec:enc}
Best-performing single-pose forecasting GCN encoders have considered two main aspects: the space-time separability of adjacency weight matrices and learning the body kinematic graph connectivity and weights. 
We detail these two aspects and empirically compare them in Table~\ref{tab:interaction}. 
Furthermore, we also consider hierarchical representations of the skeleton proposed by \cite{Dang21}, but this is not a best practice, as we determine experimentally.
Nor is it a good practice to add attention, as we discuss in this section and quantitatively evaluate in the next.

\paragraph{Space-time separability}\label{par:sep} \greencheck \\
Each graph's intra-relations are expressed through a GCN-based framework that encodes the spatiotemporal motion and the relationships between keypoints in one's skeleton~\cite{yan18, sofianos21}.
Tensor $\mathbf{X} \in \mathbb{R}^{T \times 2J \times C}$ represents a couple's skeleton pose and motion, adjacency matrices ${A}_s \in \mathbb{R}^{T \times 2J \times 2J}$ and ${A}_t \in \mathbb{R}^{2J \times T \times T}$ are responsible for learning spatial and temporal interactions respectively, as in \cite{sofianos21}.
Matrices are fully learnable, no kinematic tree is used, and the model is free to grasp the relation between body joints.
Thus, this module is formulated as follows:
\begin{equation}
    \label{eq:gcn}
    {\mathbf{Y}} = \sigma({A}_s{A}_t\mathbf{X}{W}),
\end{equation}
where $\sigma$ is an activation function and $W \in \mathbb{R}^{C \times C'}$ is a tensor of learnable weights defined as a convolution with kernel dimension $k=1$. 
Thus, it is conceptually similar to a fully connected layer. 
However, unlike the MLP design of ~\cite{guo2022back}, GCN shares the weights of $W$ across all channels.

\paragraph{Learning the graph connectivity and weights} \greencheck \\
Some works~\cite{Dang21, yan18} use inductive biases based on the human body, such as kinematic trees or specifically-devised connectivity weights. In contrast, others learn the graph adding a constraint on the optimization by spectral normalization~\cite{kipf2018neural}.
Instead, we follow what is done in the most recent work~\cite{sofianos21}: unconstrained optimization of graph edges and weights i.e., we set $A_{st}$ for nonseparable GCN and $A_s$, $A_t$ in case of space-time separable GCN as a fully learnable matrix. 
This is effectively a best practice, experimentally proven in Table \ref{tab:interaction}.
\paragraph{Attention} ~\\
A GCN model equipped with attention is also known as a Graph Attention Network (GAT)~\cite{velickovic18}. In a GAT, attention re-defines the adjacency matrix terms as a function of the node embeddings. We employ attention to encode the relation between the two actor embeddings ${B}_{h}^1$ and ${B}_{h}^2$:
\begin{equation}
    \label{eq:pain}
    {\mathbf{\mathcal{B}_{h}^1}} = {\mathcal{B}}^1 {W_{1}}, {\mathbf{\mathcal{B}_{h}^2}} = {\mathcal{B}}^2 {W_{2}},
\end{equation}
Where ${\mathcal{B}^1}, {\mathcal{B}^2} \in \mathbb{R}^{T \times J \times C}$ and $W_{1}, W_{2} \in \mathbb{R}^{C \times C}$ are learnable weights to map features in a high-dimensional space. We use these features to calculate attention weights as follows:
\begin{equation}
    \label{eq:pattention}
    {\mathbf{\eta}} = softmax\Bigl(\sigma({{\mathcal{B}}_{h}^1} {W_{3}} || ({{\mathcal{B}}_{h}^2} {W_{4}})^{\mathsf{T}})\Bigr),
\end{equation}
Where ${\mathcal{B}_{h}^1}, {\mathcal{B}_{h}^2} \in \mathbb{R}^{T \times J \times C}$, $W_{3}, W_{4}   \in \mathbb{R}^{C \times 1}$ and $\sigma$ is a LeakyRelu activation function. We apply softmax to get attention weights $\eta \in \mathbb{R}^{T \times n \times m}$ constituiting $n$ joints in $\mathcal{B}^1$ and $m$ joints in $\mathcal{B}^2$ and reweight ${B}_{h}^1$ and ${B}_{h}^2$ as follows: 
\begin{equation}
    \label{eq:paout1}
    {\mathbf{\mathcal{B}_{out}^1}} = {\mathcal{B}}_{h}^1 {\eta},{\mathbf{\mathcal{B}_{out}^2}} = {\mathcal{B}}_{h}^2 {\eta^{\mathsf{T}}},
\end{equation}
Where ${\mathcal{B}_{h}^1}, {\mathcal{B}_{h}^2} \in \mathbb{R}^{T \times J \times C}$ and $B_{out}^1$, $B_{out}^2$ are the outputs of attention module. 
We observe that in its more common use~\cite{velickovic18}, graph attention is used to estimate the interaction coefficients of the adjacency matrix $A$.
This is done by learning a function (general MLP) of two node embeddings.
By contrast, when the nodes of the graph are semantically given (body parts of a leader and follower person), one may learn the interaction coefficient (i.e., \ each term of $A$) directly, with a joint function of all nodes (not just pairs).
The direct estimation results in better performance, as shown by the experiments in Sec. \ref{sec:best_discussion}. Hence, the GCN with fully-learned parameters is selected as a best practice rather than attention.

\paragraph{Hierarchical body parts} ~\\
To the best of our knowledge, a high-level motion representation improves the prediction of human poses\cite{Li_2020_CVPR}.
\cite{Dang21} achieves this by concatenating the higher level as an extra node and hand-crafting ad-hoc neighborhoods of nodes. \\
We integrate a module within the model that enables it to decrease the number of skeleton keypoints for both bodies.
We allow the model to naturally learn aggregations between nodes by excluding artificial aggregations while shifting between hierarchies.
We employ a linear layer that learns an optimized aggregation when downscaling, and the same is done when upscaling to retrieve the original size skeleton. Although we gain a small improvement by adopting hierarchies, it becomes a limiting factor rather than a gain when combined with other best practices.

\subsection{Decoding Best Practices}\label{ssec:dec}
In earlier works, convolutions have been employed for the decoding stage \cite{gehring2017convolutional, luo2018faf, sofianos21}. However, the most recent SoA method chose a plain, fully connected layer~\cite{guo2022back}. In this section, we will analyze the two solutions, and in Sec.~\ref{sec:results}, we will show why we choose the latter.

\paragraph{Convolutional-based decoder}\label{par:conv} ~\\
In the convolutional-based decoder, convolutional layers applied to the temporal dimension are responsible for estimating the pose. It aims to forecast the subsequent frames, $t+1$ to $t+n$, given the first $t$ frames. This structure is known as Temporal Convolutional Network (TCN)~\cite{gehring2017convolutional, luo2018faf, sofianos21}.

\paragraph{FC-based decoder}  \greencheck \label{par:fc}\\
The decoder consists of a single linear layer~\cite{guo2022back} in charge of mapping the observed $T$ frames to the predicted $N$.

\subsection{Novel Adjacency Matrix Initialization}\label{ssec:init}

We propose a novel initialization methodology, aiming to preserve variance during the forward pass, which matches the preservation of gradients in the backward.
Since over several layers a non-unit variance results in vanishing or exploding signals, and neither of those is good for training, as they stall the gradient, we aim to preserve the variance.
To do that, under the assumption of a neural network consisting of only linear layers and linear activation functions, \cite{glorot10} proposes to estimate the standard deviation by considering the number of neurons in both the current and previous layer.

It is particularly relevant for our model because it comprises 8 layers while GCNs are often shallow~\cite{kipf17}.
We propose to randomly initialize the fully learnable matrices $A_s$, $A_t$, and $W$ according to a uniform distribution, whose bounds are defined in such a way that considers both the number of graph nodes and the number of timeframes.

Convolutions on graphs that adopt a normalized adjacency matrix~\cite{kipf17, tong20} use a well-known graph and do not let all nodes interact with each other.
Furthermore, normalization avoids vanishing and exploding gradient, yet it limits the performance and, in the end, fully-learnable yields the best performances~\cite{sofianos21, sampieri22}.
Here is the importance of randomly initializing an \textit{ad hoc} fully learnable adjacency matrix, avoiding exploding or vanishing gradients.
The response from the Separable GCN at layer $l$, according to Eq.~\eqref{eq:gcn}, is
\begin{equation}
    \label{eq:gcn:l}
    \mathbf{X}^{l+1} = \sigma \left( A_s^l A_t^l \mathbf{X}^l W^l \right), \quad \forall l.
\end{equation}
Let's assume matrices $A_s$, $A_t$, and $W$ to be independent, have zero mean~\cite{glorot10, he15} and uniformly distributed.
To constrain variance, hence stabilize training and avoid exploding or vanishing gradient, constraining the variance of the output product of $n^l$ neurons at layer $l$ times $W$ to $1$~\cite{he15} is a sufficient condition, i.e.,
\begin{equation}
    \label{eq:init:constraint:w}
    \frac{1}{k} n^l Var[W^l] = 1, \quad \forall l,
\end{equation}
where $k = 2$ in the case of Re-LU activations, which are asymmetric~\cite{he15} (while $k=1$ for symmetric activations such as the \texttt{tanh}).
For the spatial matrix, rather than the number of neurons $n^l$, we consider the number of nodes $v$, which $A_s$ integrates
\begin{align}
    \label{eq:init:constraint:As}
    \frac{1}{k}(n_v^l) Var[A_s^l] &= 1, \quad \forall l.
\end{align}
Similarly, we consider $t$ time frames to initialize the temporal matrix $A_t$,
\begin{align}
    \label{eq:init:constraint:At}
    \frac{1}{k}(n_t^l) Var[A_t^l] &= 1, \quad \forall l.
\end{align}

When initializing $W$ with a zero-mean uniform distribution, the constraint of Eq.~\eqref{eq:init:constraint:w} yields the following distribution for the initialization:
\begin{equation}
    \label{eq:init:W}
    W^l \sim U \left[ - \sqrt{\frac{k}{n^l}}, \sqrt{\frac{k}{n^l}} \right], \quad \forall l.
\end{equation}
The spatial and temporal matrix constraints of Eqs.~\eqref{eq:init:constraint:As} and \eqref{eq:init:constraint:At} translate to the following initializing distributions for $A_s$ and $A_t$ respectively:
\begin{align}
    \label{eq:init:As}
    A_s^l &\sim U \left[ - \sqrt{\frac{k}{n_v^l}}, \sqrt{\frac{k}{n_v^l}} \right], \\
    \label{eq:init:At}
    A_t^l &\sim U \left[ - \sqrt{\frac{k}{n_t^l}}, \sqrt{\frac{k}{n_t^l}} \right], \quad \forall l.
\end{align}

\begin{table*}[!ht]
\centering
\renewcommand{\arraystretch}{1.2}
\resizebox{\textwidth}{!}{
\begin{tabular}{l|c@{\hspace{0.3\tabcolsep}}c@{\hspace{0.3\tabcolsep}}c@{\hspace{0.3\tabcolsep}}c|c@{\hspace{0.3\tabcolsep}}c@{\hspace{0.3\tabcolsep}}c@{\hspace{0.3\tabcolsep}}c|c@{\hspace{0.3\tabcolsep}}c@{\hspace{0.3\tabcolsep}}c@{\hspace{0.3\tabcolsep}}c|c@{\hspace{0.3\tabcolsep}}c@{\hspace{0.3\tabcolsep}}c@{\hspace{0.3\tabcolsep}}c|c@{\hspace{0.3\tabcolsep}}c@{\hspace{0.3\tabcolsep}}c@{\hspace{0.3\tabcolsep}}c|c@{\hspace{0.3\tabcolsep}}c@{\hspace{0.3\tabcolsep}}c@{\hspace{0.3\tabcolsep}}c|c@{\hspace{0.3\tabcolsep}}c@{\hspace{0.3\tabcolsep}}c@{\hspace{0.3\tabcolsep}}c|c@{\hspace{0.3\tabcolsep}}c@{\hspace{0.3\tabcolsep}}c@{\hspace{0.3\tabcolsep}}c}
\hline
 \multicolumn{1}{c|}{\textit{Action}}                      & \multicolumn{4}{c|}{A1}                                                    & \multicolumn{4}{c|}{A2}                                            & \multicolumn{4}{c|}{A3}                                                                  & \multicolumn{4}{c|}{A4}                                               & \multicolumn{4}{c|}{A5}                                                      & \multicolumn{4}{c|}{A6}                                                   & \multicolumn{4}{c|}{A7}                                                  & \multicolumn{4}{c}{Average $\downarrow$}                                                       \\ \hline
\multicolumn{1}{c|}{\textit{Time (msec)}} & {  \textit{200}} & {  \textit{400}} & {  \textit{600}} & {  \textit{1000}} & {  \textit{200}} & {  \textit{400}} & {  \textit{600}} & {  \textit{1000}} & {  \textit{200}} & {  \textit{400}} & {  \textit{600}} & {  \textit{1000}}               & {  \textit{200}} & {  \textit{400}} & {  \textit{600}} & {  \textit{1000}} & {  \textit{200}} & {  \textit{400}} & {  \textit{600}} & {  \textit{1000}} & {  \textit{200}} & {  \textit{400}} & {  \textit{600}} & {  \textit{1000}} & {  \textit{200}} & {  \textit{400}} & {  \textit{600}} & {  \textit{1000}} & {  \textit{200}} & {  \textit{400}} & {  \textit{600}} & {  \textit{1000}} \\ \hline \hline
LTD~\cite{mao19ltd}                                          & 70                 & 125                & 157                & 189                 & 131                & 242                & 321                & 426                 & 102                & 194                & 260                & \multicolumn{1}{c|}{357}          & 62                 & 117                & 155                & 197                 & 72                 & 131                & 173                & 231                 & 81                 & 151                & 200                & 280                 & 112                & 223                & 315                & 442                 & 90                 & 169                & 226                & 303                 \\
 HisRep~\cite{mao20his}                                           & 52                 & 103                & 139                & 188                 & 96                 & 186                & 256                & 349                 & 57                 & 118                & 167                & \multicolumn{1}{c|}{240}          & 45                 & 93                 & 131                & 180                 & 51                 & 105                & 149                & 214                 & 61                 & 125                & 176                & 252                 & 71                 & 150                & 222                & 333                 & 62                 & 126                & 177                & 251                 \\
 MSR-GCN~\cite{Dang21}                                         & 56                 & 100                & 132                & 175                 & 102                & 187                & 256                & 365                 & 65                 & 120                & 166                & \multicolumn{1}{c|}{244}          & 50                 & 95                 & 127                & 172                 & 54                 & 100                & 138                & 202                 & 70                 & 132                & 182                & 258                 & 82                 & 154                & 218                & 321                 & 69                 & 127                & 174                & 248                 \\
 MRT~\cite{wang21}                                         &  50 & 98 & 134 & 188 & 79 & 155 & 212 & 307 & 53 & 106 & 152 & \multicolumn{1}{c|}{229} & 47 & 95 & 131 & 185 & 52 & 105 & 149 & 215 & 58 & 118 & 166 & 242 & 65 & 136 & 199 & 299 & 58 & 116 & 163 & 238                  \\
siMLPe~\cite{guo2022back}                                  & 49                 & 102                 & 137                & 177                 & 88                 & 180                & 244                & 336                 & 57                 & 122                & 174                & \multicolumn{1}{c|}{254}          & 45                 & 100                 & 137                & 182                 & 50                 & 103                 & 144                & 206                 & 59                 & 126                & 175                & 250                 & 77                 & 164                & 134                & 348                 & 60                 & 128                & 178                & 250                 \\ 
XIA~\cite{guo21}                                  & 49                 & 98                 & 140                & 192                 & 84                 & 166                & 234                & 346                 & 51                 & 105                & 154                & \multicolumn{1}{c|}{234}          & 41                 & 84                 & 120                & 161                 & 43                 & 90                 & 132                & 197                 & 55                 & 113                & 163                & 242                 & 62                 & 130                & 192                & 291                 & 55                 & 112                & 162                & 238                 \\ 

Ours                               & \textbf{34}        & \textbf{71}        & \textbf{105}       & \textbf{159}        & \textbf{56}        & \textbf{121}       & \textbf{181}       & \textbf{292}        & \textbf{36}        & \textbf{78}        & \textbf{118}       & \multicolumn{1}{c|}{\textbf{195}} & \textbf{30}        & \textbf{66}        & \textbf{98}       & \textbf{145}        & \textbf{35}        & \textbf{74}        & \textbf{113}       & \textbf{171}        & \textbf{41}        & \textbf{88}        & \textbf{129}       & \textbf{193}        & \textbf{47}        & \textbf{108}       & \textbf{166}       & \textbf{261}        & \textbf{39}        & \textbf{86}        & \textbf{129}       & \textbf{202}        \\ \hline 

\end{tabular}}
\caption{Results in millimeters for ExPI Common actions split. Our model achieves state-of-the-art results in all actions considered, at each predicted time instant. }
\label{tab:expi}
\end{table*}
\begin{table*}[!htb]
\centering
% \addtolength{\tabcolsep}{-2pt}

\renewcommand{\arraystretch}{1.2}
\resizebox{\textwidth}{!}{
\begin{tabular}{l|c@{\hspace{0.7\tabcolsep}}c@{\hspace{0.7\tabcolsep}}c|
c@{\hspace{0.7\tabcolsep}}c@{\hspace{0.7\tabcolsep}}c|
c@{\hspace{0.7\tabcolsep}}c@{\hspace{0.7\tabcolsep}}c|
c@{\hspace{0.7\tabcolsep}}c@{\hspace{0.7\tabcolsep}}c|c@{\hspace{0.7\tabcolsep}}c@{\hspace{0.7\tabcolsep}}c|c@{\hspace{0.7\tabcolsep}}c@{\hspace{0.7\tabcolsep}}c|c@{\hspace{0.7\tabcolsep}}c@{\hspace{0.7\tabcolsep}}c|c@{\hspace{0.7\tabcolsep}}c@{\hspace{0.7\tabcolsep}}c|c@{\hspace{0.7\tabcolsep}}c@{\hspace{0.7\tabcolsep}}c|c@{\hspace{0.7\tabcolsep}}c@{\hspace{0.7\tabcolsep}}c}
\hline
 \multicolumn{1}{c|}{\textit{Action}}                     & \multicolumn{3}{c|}{A8}                     & \multicolumn{3}{c|}{A9}                     & \multicolumn{3}{c|}{A10}                                      & \multicolumn{3}{c|}{A11}                                      & \multicolumn{3}{c|}{A12}                                      & \multicolumn{3}{c|}{A13}                                      & \multicolumn{3}{c|}{A14}                                      & \multicolumn{3}{c|}{A15}                                      & \multicolumn{3}{c|}{A16}                                      & \multicolumn{3}{c}{Average $\downarrow$}                                  \\ \hline 
 \multicolumn{1}{c|}{\textit{Time (msec)}} & {   \textit{400}} & {   \textit{600}} & {   \textit{800}} & {   \textit{400}} & {   \textit{600}} & {   \textit{800}} & {   \textit{400}} & {   \textit{600}} & {   \textit{800}} & {   \textit{400}} & {   \textit{600}} & {   \textit{800}} & {   \textit{400}} & {   \textit{600}} & {   \textit{800}} & {   \textit{400}} & {   \textit{600}} & {   \textit{800}} & {   \textit{400}} & {   \textit{600}} & {   \textit{800}} & {   \textit{400}} & {   \textit{600}} & {   \textit{800}} & {   \textit{400}} & {   \textit{600}} & {   \textit{800}} & {   \textit{400}} & {   \textit{600}} & {   \textit{800}} \\ \hline \hline
LTD~\cite{mao19ltd}                                              & 252          & 333          & 387           & 174          & 228          & 268           & 139                & 184                & 217                 & 239                & 324                & 394                 & 175                & 226                & 259                 & 148                & 191                & 220                 & 176                & 240                & 286                 & 143                & 178                & 192                 & 146                & 193                & 226                 & 177                & 233                & 272                 \\
HisRep~\cite{mao20his}                                           & 157          & 219          & 257           & 134          & 190          & 233           & 96                 & 146                & 187                 & 195                & 283                & 358                 & 121                & 169                & 206                 & 92                 & 129                & \textbf{160}                 & 129                & 193                & 245                 & 80                 & 104                & 121                 & 112                & 154                & 187                 & 124                & 176                & 218                 \\
MSR-GCN~\cite{Dang21}                                          & 177          & 239          & 295           & 143          & 179          & 213           & 157                & 222                & 281                 & 230                & 289                & 335                 & 188                & 245                & 290                 & 148                & 198                & 248                 & 234                & 319                & 384                 & 176                & 232                & 278                 & 162                & 218                & 266                 & 179                & 238                & 288                 \\
 MRT~\cite{wang21}                                          &  170 & 231 & 308 & 145 & 199 & 270 & 141 & 245 & 338 & 225 & 327 & 481 & 131 & 180 & 253 & 120 & 169 & 238 & 165 & 229 & 322 & 110 & 151 & 209 & 105 & 144 & 201 & 146 & 205 & 291                  \\
siMLPe~\cite{guo2022back}                                  & 165          & 220          & 258           & 137          & 198          & 246           & 104                 & 154                & 198                 & 210                & 301                & 432                 & 114      & 156       & 187        & 94        & 132       & 160       & 140       & 204       & 255        & 91        & 119       & 138        & 120                & 166                & 204                 & 131                & 183                & 225                 \\ 
XIA~\cite{guo21}                                  & 156          & 216          & 256           & 126          & 175          & 213           & 96                 & 152                & 205                 & 191                & 287                & 377                 & 118      & 165       & 203        & 91        & 129       & 162       & 122       & 183       & 232        & 81        & \textbf{107}       & \textbf{128}        & 106                & 150                & 185                 & 121                & 174                & 218                 \\ 

Ours                               & \textbf{113} & \textbf{164} & \textbf{203}  & \textbf{114}  & \textbf{167} & \textbf{209}  & \textbf{85}        & \textbf{136}       & \textbf{183}        & \textbf{153}       & \textbf{231}       & \textbf{304}        & \textbf{100}                & \textbf{148}                & \textbf{188}                & \textbf{82}                 & \textbf{125}                & 162                 & \textbf{91}                & \textbf{138}                & \textbf{179}                 & \textbf{79}                & 109               & 132                 & \textbf{85}        & \textbf{124}       & \textbf{156}        & \textbf{100}       & \textbf{149}       & \textbf{191}        \\ \hline
\end{tabular}}
\caption{Results in millimeters for ExPI Unseen actions split. On average, we outperform the baseline considered over short and long time horizons.  } 
\label{tab:unseen}
\end{table*}

%%%%%%%%% RESULTS
\section{Experiments}
\label{sec:results}

We thoroughly evaluate the proposed best practices on the most recent and challenging 2-body pose forecasting dataset ExPI~\cite{guo21}, comparing against the SoA and the best single-pose forecasting techniques adapted to the task. The selected best practices also perform on par with the SoA in single-pose forecasting on the established Human3.6M dataset~\cite{h36m}.

\subsection{Benchmark and baselines}

\paragraph{Datasets.} The dataset used for multi-body pose forecasting, ExPI~\cite{guo21}, is a collection of two different dancing pairs performing Lindy Hop sessions, dubbed ``extreme human interaction'' by the authors~\cite{guo21}.
Data were collected in a multi-camera platform with 68 synchronized and calibrated RGB cameras and a motion capture system with 20 mocap cameras. The missing points were manually fixed to ensure good data quality. 
ExPI contains 115 sequences at 25 fps with 18 body joints for each of the two persons involved.
These agents are grouped in two couples, dubbed $(\mathcal{A}^1_{c}, \mathcal{A}^2_{c})$, which perform 16 different actions. Actions A1 to A7 are common to both couples; A8 to A13 performed only by $\mathcal{A}^1_{c}$ and A14-A16 by $\mathcal{A}^2_{c}$.
Based on this, ExPI provides three different splits to test the model on:
\begin{itemize}
    \item \textbf{Common.} Training and test set are composed only of actions performed by both couples. The ones belonging to $\mathcal{A}^2_{c}$ define the train set, and $\mathcal{A}^1_{c}$'s the test set.
    \item \textbf{Unseen.} Differently from the previous one, this split has common actions to both $\mathcal{A}^1_{c}$ and $\mathcal{A}^2_{c}$ as the train set and couple-specific actions as the test one. 
    This subset allows us to test for generalization.
    \item \textbf{Single.}
    In this split, a single action from couple $\mathcal{A}^2_{c}$ is used as a train set, and the same action from couple $\mathcal{A}^1_{c}$ as the test set. 
    It allows testing how the model generalizes to a new couple for each action.
\end{itemize}

We also test on Human3.6M~\cite{h36m}, an established dataset for single-person pose forecasting.
It consists of a total of 3.6 million poses, acquired at 25 fps, depicting seven actors performing 15-day real-life actions, e.g., walking, sitting, and talking on the phone.
Following~\cite{mao21multi, mao20his, Dang21}, we train on subjects S1, S6, S7, S8, S9, we use S11 for validation, and S5 for testing.

\paragraph{Evaluation metrics.}
We validate performance by the \textit{Mean per joint position Error}, defined as the MPJPE \cite{h36m, mao19ltd} and renamed as JME in ~\cite{guo21}  at a future frame $t$:
\begin{equation}
    L_{\text{JME}} = L_{\text{MPJPE}} = \frac{1}{V}\sum^V_{v=1} ||\hat{x}_{vt} - x_{vt}||_2,
    \label{jme}
\end{equation}

where $\hat{x}_{vt}$ and $x_{vt}$ are the 3-dimensional vectors of a target joint and the ground truth, respectively.
For the joint evaluation of the 2-body position error, the two body poses are normalized into the same reference system. In this work, we keep the MPJPE notation. 

\paragraph{Baselines.} 
 We select the latest and best-performing single-body pose forecasting models, and we adapt them to predict the motion of two people.
XIA-Transformer~\cite{guo21} is the only 2-body pose forecasting method in the literature. XIA uses a transformer to encode skeleton features and model the body-body interaction via attention. We consider \cite{wang21} the only multi-body model based on a Transformer architecture.
Due to the lack of multi-body pose forecasting models, we also compare them to single ones. 
LTD~\cite{mao19ltd} consists of a cascade of GCN blocks acting on frequencies, and its extension, HisRep~\cite{mao20his}, inserts a motion attention mechanism based on DCT coefficients operating on sub-sequences of the input. 
MSR-GCN~\cite{Dang21} is a hierarchical GCN-based technique that applies multi-scale aggregations, so coarser scales represent groups of body joints and coarser motion. 
In Table \ref{tab:h36m} we compare ourselves, again,  to LTD~\cite{mao19ltd}, HisRep~\cite{mao20his} and MSR-GCN~\cite{Dang21} and, additionally, on two recent single-body models.
SeS-GCN~\cite{sampieri22} adopts an all-separable GCN with a teacher-student approach, and the SoA~\cite{guo2022back}, which consists of MLPs encoding spatial and temporal relationships. 

\begin{table*}[!htb]
\centering
\renewcommand{\arraystretch}{1.2}
\resizebox{\textwidth}{!}{
\begin{tabular}{l|c@{\hspace{0.7\tabcolsep}}c@{\hspace{0.7\tabcolsep}}c@{\hspace{0.7\tabcolsep}}c|c@{\hspace{0.7\tabcolsep}}c@{\hspace{0.7\tabcolsep}}c@{\hspace{0.7\tabcolsep}}c|c@{\hspace{0.7\tabcolsep}}c@{\hspace{0.7\tabcolsep}}c@{\hspace{0.7\tabcolsep}}c|c@{\hspace{0.7\tabcolsep}}c@{\hspace{0.7\tabcolsep}}c@{\hspace{0.7\tabcolsep}}c|c@{\hspace{0.7\tabcolsep}}c@{\hspace{0.7\tabcolsep}}c@{\hspace{0.7\tabcolsep}}c|c@{\hspace{0.7\tabcolsep}}c@{\hspace{0.7\tabcolsep}}c@{\hspace{0.7\tabcolsep}}c|c@{\hspace{0.7\tabcolsep}}c@{\hspace{0.7\tabcolsep}}c@{\hspace{0.7\tabcolsep}}c}
\hline
\multicolumn{1}{c|}{\textit{Action}}             & \multicolumn{4}{c|}{A1}                         & \multicolumn{4}{c|}{A2}                  & \multicolumn{4}{c|}{A3}                          & \multicolumn{4}{c|}{A4}                    & \multicolumn{4}{c|}{A5}                           & \multicolumn{4}{c|}{A6}                         & \multicolumn{4}{c}{A7}                        \\ \hline  
\multicolumn{1}{c|}{\textit{Time (msec)}} & \textit{200}         & \textit{400}         & \textit{600}          & \textit{1000}         & \textit{200}         & \textit{400}          & \textit{600}          & \textit{1000}         & \textit{200}         & \textit{400}          & \textit{600}          & \textit{1000}         & \textit{200}         & \textit{400}         & \textit{600}          & \textit{1000}         & \textit{200}         & \textit{400}         & \textit{600}          & \textit{1000}         & \textit{200}         & \textit{400}          & \textit{600}          & \textit{1000}         & \textit{200}         & \textit{400}          & \textit{600}          & \textit{1000}         \\ \hline \hline
LTD~\cite{mao19ltd}                                     & 70          & 126         & 155          & 183          & 131         & 243          & 312          & 415          & 102         & 194          & 252          & 338          & 62          & 117         & 153          & 203          & 71          & 131         & 171          & 231          & 81          & 151          & 199          & 299          & 112         & 223          & 306          & 411          \\
HisRep~\cite{mao20his}                                  & 66          & 118         & 153          & 190          & 128         & 231          & 308          & 417          & 74          & 143          & 205          & 295          & 64          & 120         & 159          & 191          & 63          & 121         & 166          & 227          & 90          & 168          & 232          & 312          & 88          & 166          & 232          & 332          \\
MSR-GCN~\cite{Dang21}                                 & 64          & 108         & 136          & \textbf{170}          & 119         & 210          & 282          & 385          & 79          & 144          & 189          & 265          & 59          & 103         & 134          & 173          & 65          & 118         & 162          & 225          & 86          & 151          & 201          & 283          & 96          & 178          & 255          & 362          \\
 MRT~\cite{wang21}               &  63 & 120 & 160 & 218 & 97 & 190 & 249 & 346 & 77 & 148 & 193 & 240 & 51 & 102 & 139 & 186 & 61 & 118 & 163 & 226 & 58 & 115 & 151 & 198 & 82 & 172 & 244 & 340  \\
siMLPe~\cite{guo2022back}                         & 60          & 113         & 145          & 200          & 104         & 202          & 268          & 373          & 76 & 150 & 205 & 305          & 58          & 110         & 151          & 203          & 64          & 123         & 163          & 218          & 76          & 152          & 207          & 277          & 93 & 180 & 254 & 341 \\ 
XIA~\cite{guo21}                         & 64          & 120         & 160          & 199          & 109         & 200          & 275          & 381          & 59 & 117 & 174 & 277          & 60          & 116         & 162          & 209          & 53          & 106         & 152          & 221          & 65          & 122          & 166          & 223          & 74 & 144 & \textbf{203} & \textbf{301} \\ 

Ours                            & \textbf{52} & \textbf{94} & \textbf{128} & 179 & \textbf{89} & \textbf{176} & \textbf{242} & \textbf{329} & \textbf{42}          & \textbf{90}          & \textbf{129}          & \textbf{200} & \textbf{49} & \textbf{96} & \textbf{134} & \textbf{185} & \textbf{48} & \textbf{99} & \textbf{140} & \textbf{196} & \textbf{52} & \textbf{105} & \textbf{144} & \textbf{198} & \textbf{68} & \textbf{140}          & 204         & 305          \\ \hline
\end{tabular}}
\caption{Results in millimeters for ExPI Single actions split. We outperform in 6 out of 7 stocks all baselines considered according to the MPJPE metric. For the other stocks our model is comparable with the current state of the art. }
\label{tab:single}
\end{table*}

\subsection{Evaluation of human pose forecasting} 
We evaluate our model quantitatively and qualitatively on ExPI's \cite{guo21} provided splits. We further test our model's generalization power on the single-body dataset Human3.6M.

\paragraph{ExPI Common Actions.} \label{par:common}
Table~\ref{tab:expi} shows the results obtained from our best model with our selected best practices. 
These outperform every tested method by a large margin, both the SoA single-person and the SoA 2-body pose forecasting techniques.
The overall mean improvement is 22\% over all actions and all time horizons. In particular, on all actions, the improvement for short-term future predictions (200 msec) is 29\% and 15\% for the long-term.

\paragraph{ExPI Unseen Actions.} 
Table \ref{tab:unseen}  also showcases improvements using the proposed best practices. On average, across all forecasting horizons, the improvement is 14\%. 

\paragraph{ExPI Single Actions.} In Table \ref{tab:single}, also for the case of single actions, the best practices report an average improvement of 14.2\%. They outperform all other tested techniques in 6 (out of 7) actions at all predicted time horizons. It confirms the generalization of our model to new people.

\begin{figure}[htbp]
    \centering
    \includegraphics[scale=0.19]{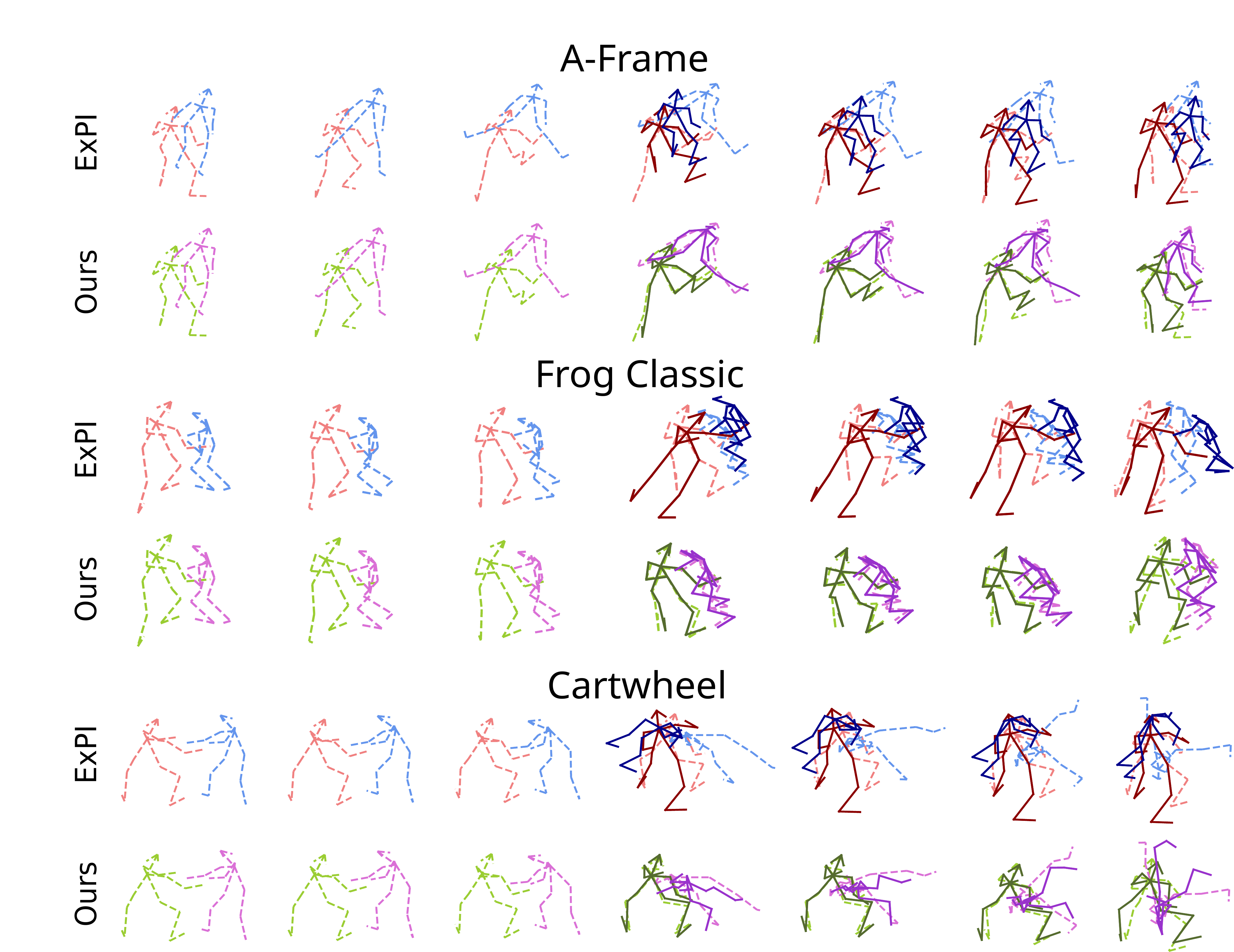}

    \caption{Visual comparison of our proposed best-practice  model (\emph{Ours}) against ExPI~\cite{guo21}. The first three columns are observed, and the last four are predicted poses.
    Light-colored and dashed skeletons are GT, and darker and solid ones are predictions. Note the improved larger-displacement motions (\textit{Cartwheel}).
    %A1 - A-frame, A4 - Frog classic, A7 - Cartwheel
    }
     \label{fig:poses-viz}
\end{figure}

\vspace{-.3cm}
\paragraph{ExPI qualitative.}
In Fig.~\ref{fig:poses-viz}, the current SoA, ExPI~\cite{guo21}, is compared against the best-practice model (\emph{Ours}), qualitatively. The first three columns depict observations; the following four are future motion predictions.
The light-colored pictograms represent ground-truth motion.
The best practices provide, in general, better predictions. Best improvements are observed in the case of large motion displacements, cf.\ the last two rows, action ``Cartwheel''.

\begin{table}[!h]
\centering

\begin{tabular}{lcccc}
\hline
                                          & \multicolumn{4}{c}{MPJPE $\downarrow$}                                                       \\ \cline{2-5}
\multicolumn{1}{l}{\textit{Time Horizon (msec)}} & {\textit{160}} & {\textit{400}} & {\textit{560}} & {\textit{1000}} \\ \hline \hline
\multicolumn{1}{l|}{LTD~\cite{mao19ltd}}                 & 23.4               & 58.9               & 78.3               & 114.0               \\
\multicolumn{1}{l|}{HisRep~\cite{mao20his}}              & 22.6                 & 58.3               & 77.3               & 112.1               \\
\multicolumn{1}{l|}{MSR-GCN~\cite{Dang21}}             & 25.5               & 63.3               & 81.1               & 114.1               \\
% \multicolumn{1}{l|}{ST-DGCN~\cite{mao21multi}}             & 23.1               & 57.9               & 76.3               & 109.7               \\
\multicolumn{1}{l|}{SeS-GCN ~\cite{sampieri22}}             & 29.0           & 64.0               & 84.4               & 113.9               \\
\multicolumn{1}{l|}{siMLPe~\cite{guo2022back}}             & \textbf{21.7}               & \textbf{57.3}               & \textbf{75.7}               & \textbf{109.4}               \\ \hline
\multicolumn{1}{l|}{Ours w/o init.}  & 27.3              & 64.6               & 83.1               & 116.3               \\ 
\multicolumn{1}{l|}{Ours}  & 26.8               & 63.1               & 81.1               & 113.2               \\ \hline
\end{tabular}
\caption{Error in millimeters on Human3.6M dataset. We show how our method adapted to single-person human pose forecasting is comparable with the best-performing techniques on average.}
\label{tab:h36m}
\end{table}
\vspace{-.5cm}

\begin{table*}[t]
% \rowcolors{3}{gray!12}{white}
\centering
\resizebox{1\textwidth}{!}{%
% \begin{tabular}{c|c|cccc|c|cccc|c} 
\begin{tabular}{c|ccccccccccccc} 
\hline
\multicolumn{1}{c}{~} & \multicolumn{1}{c|}{\textbf{Model}} & \multicolumn{1}{c|}{\textbf{Input Repr.}} & \multicolumn{5}{c|}{\textbf{Encoding}} & \multicolumn{1}{c|}{\textbf{Decoding}}  &   \multicolumn{4}{c|}{\textit{\textbf{MPJPE} $\downarrow$}} & \multicolumn{1}{c}{\textbf{Param. }$\downarrow$} \\
% \hline
% \cline{8-11}
% \cline{4-6}
 \multicolumn{1}{c}{~} & \multicolumn{1}{c|}{~} & \multicolumn{1}{c|}{Freq. Enc. \color{ForestGreen}\cmark} &  \multicolumn{1}{c}{Learn. \color{ForestGreen}\cmark} & \multicolumn{1}{c}{Sep. \color{ForestGreen}\cmark} & Init. \color{ForestGreen}\cmark & Att. & \multicolumn{1}{c|}{Hier.} & \multicolumn{1}{c|}{FC \color{ForestGreen}\cmark} & \textit{200}	& \textit{400} & \textit{600}	& \multicolumn{1}{c|}{\textit{1000}} & (M)\\ \hline \hline
% Model & \multicolumn{1}{c}{Freq. Emb.} & \multicolumn{1}{c}{Init.} & Sep. & Attention & \multicolumn{1}{c}{Hier.} & \multicolumn{1}{c}{FC decoder} & \textit{200}	& \textit{400} & \textit{600}	& \textit{1000}  \\ \hline \hline

1 & \cite{guo21} &  \cmark & \cmark & & & {\cmark} &  & & 55 & 112 & 162 & 238 & 8.5  \\
\hline
% MLP \cite{guo2022back} & &  &  & & & {\cmark} & - & - & - & - & - \\
2 & Space-time GCN & ~ & \cmark & ~ & ~ & ~ & ~ & ~ & 108 & 152 & 255 & 379 & 1.08 \\
3 & \textit{(kin. tree)} & ~ & ~ & \cmark & ~ & ~ & ~ & ~ & 81 & 129 & 183 & 260 & 0.18 \\
4 & ~ & ~ & \cmark & \cmark & ~ & ~ & ~ & ~ & 55 & 112 & 156 & 224 & 0.18 \\
\hline
5 & Input repr. practice & \cmark & \cmark & \cmark & ~ & ~ & ~ & ~ & 41 & 88 & 135 & 219 & 0.18 \\
\hline
6 & ~ & ~ & \cmark & \cmark & \cmark & ~ & ~ & ~ & 53 & 106 & 148 & 216 & 0.18 \\
7 & Encoder practices & ~ & \cmark & \cmark & ~ & {\cmark}$^{\dagger}$ & ~ & ~ & 55 & 112 & 157 & 228 & 9.9 \\
8 & ~ & ~ & \cmark & \cmark & ~ & ~ & \cmark & ~ & 51 & 104 & 148 & 223 & 0.18 \\
\hline
9 & Decoder practices & ~ & \cmark & \cmark & ~ & ~ & ~ & \cmark & 51 & 104 & 145 & 212 & \textbf{0.17} \\
\hline
10 & ~ & \cmark & \cmark & \cmark & ~ & ~ & ~ & \cmark & 41 & 89 & 133 & 208 & \textbf{0.17} \\
11 & ~ & \cmark & \cmark & \cmark & ~ & ~ & \cmark & \cmark & 51 & 104 & 146 & 217 & \textbf{0.17} \\
 % & {\cmark} & {\cmark} & {\cmark} & {\cmark} &  & {\cmark} & 40 & 90 & 136 & 212 & 9.8 \\
12 & \textbf{Best model} & \cmark & \cmark & \cmark & \cmark & ~ & ~ & \cmark & \textbf{39} & \textbf{86} & \textbf{129} & \textbf{202} & \textbf{0.17} \\
\hline
\end{tabular}}

\caption{Combinations of best practices. From left to right, we have frequency encoding, fully learnable connections, Space-time separability, initialization, attention mechanism, hierarchy, fully connected layer as a decoder. $\dagger$: we implement a Graph Attention Network (GAT) tailored for GCNs, similar in spirit to \cite{guo21} designed for transformers. 
}
\label{tab:interaction}

\end{table*}

%\vspace{-.5cm}

\vspace{-.2cm}
\paragraph{Evaluation of single-person pose forecasting.}

We test how the 2-body best practices transfer back to single-person pose forecasting for a sanity check.
In Table~\ref{tab:h36m}, observe that the best practices (\emph{Ours}) yield results within a small margin compared to SoA.
Note that, for the sake of this experiment, we just run the 2-body best-practice model \emph{as is}. Without any hyper-parameter tuning. Furthermore, the initialization gives an overall 2.4$\%$ over the counterpart model that does not use it.

\subsection{Evaluation of Best Practices}\label{sec:best_discussion}

In this section, we refer to Table~\ref{tab:interaction} and thoroughly assess each selected practice. First, we select a baseline GCN model. Secondly, we assess each practice's performance, added as a standalone extension. Thirdly, we integrate practices. Best practices are assessed based on their standalone performance improvement and complementarity. Finally, in Table~\ref{tab:init}, we evaluate the impact of the proposed initialization in more detail.

\paragraph{Baseline selection.}
We first select a baseline model on which we test each best practice. We identify three possible GCN-based encoder architectures:
\begin{itemize}
    \item Space-time GCN~\cite{yan18}: this is a plain GCN model $\sigma(AXW)$ with learnable A (learnable connectivity and graph weights)
    \item Space-time separable GCN with learnable kinematic tree: inspired by \cite{yan18} and \cite{sofianos21} to factorize the adjacency matrix into two spatial and temporal learnable matrices, whereby the spatial connectivity is constrained to the kinematic tree
    \item Space-time separable GCN with fully-learnable connections: lastly, we evaluate a space-time separable GCN with fully-learnable adjacencies matrices taking inspiration from~\cite{sofianos21}. 
\end{itemize}

As shown in Table~\ref{tab:interaction}, the space-time GCN with separability (row 3) has an overall decrease of error by 25\% compared to the base GCN (row 2). A considerable additional performance boost (18\% over all frames) s also given by using the separability and fully-learnable connections (row 4) instead of limiting the learning procedure on the kinematic tree. The simple space-time separable GCN already outperforms XIA~\cite{guo21} while having a fraction of the parameters, although XIA includes DCT representations and attention. Thus GCN with separability and fully-learnable connections is a good baseline to build upon.

\paragraph{Standalone best practices.}
Table~\ref{tab:interaction} shows input representation (row 5), encoding (rows 6-8), and decoding practices (row 9). When considering the input representation and decoding techniques, DCT, and fully connected (FC) layer as decoder, it is clear that both have a considerable impact. The DCT provides a significant boost in short-term predictions, up to 25\%, while the FC-based decoder offers a more substantial increase in long-term predictions, up to 7\% against TCN (when the box is not \cmark ). Regarding the encoder practices, the novel initialization procedure and a hierarchical architecture improve the chosen baseline by 5\% and 4\%, respectively. On the other hand, using the attention technique did not lead to any gain in performance and is hence not considered a best practice.

\paragraph{Integrated best practices.}
Rows 10-12 in Table~\ref{tab:interaction} refers to the combinations of techniques that performed best independently.

Integrating the input representation using DCT coefficients and the FC-based decoder indicates how these two methods can be used in addition to the standard method. Secondly, we include a Graph Attention Network as explained in Sec. \ref{ssec:enc} to account for the interaction. The performance does not benefit from it, and the number of parameters is considerably higher.
Lastly, a hierarchical structure lowers performance when combined with other practices, so we do not consider it a best practice. Our proposed initialization improves our best practice model by another 3.5\%.

\paragraph{Impact of initialization.}\label{ssec:init_discussion}
Table~\ref{tab:init} shows the average of multiple runs for different initialization methods and the corresponding standard deviation. We compare our strategy with the Uniform sampling and the two established methodologies of \cite{glorot10}, and \cite{he15}. Our proposed initialization exceeds or is on par with the others on average, having more than 2.6\% improvement over uniform sampling over the longer time horizon. Note also the lower standard deviation of performance for our proposed technique, especially for the most challenging long-term prediction horizon (at least 2x lower), which we interpret as improved stability.

\begin{table}[!htb]
\centering
\resizebox{0.47\textwidth}{!}{
\begin{tabular}{lcccc}
\hline
                                          & \multicolumn{4}{c}{MPJPE $\downarrow$}                                                       \\ \cline{2-5}
\multicolumn{1}{l}{\textit{Time Horizon (msec)}} & {\textit{200}} & {\textit{400}} & {\textit{600}} & {\textit{1000}} \\ \hline \hline
\multicolumn{1}{l|}{Uniform}               & 39.7 \small{$\pm$0.7}            & 87.6 \small{$\pm$0.7}             & 132.2 \small{$\pm$0.5}              & 207.7 \small{$\pm$1.1}              \\
\multicolumn{1}{l|}{Glorot et al.\cite{glorot10}}                 & 40.3 \small{$\pm$0.1}               & 89.4 \small{$\pm$1.2}               & 134.3 \small{$\pm$1.5}               & 207.9 \small{$\pm$1.8}              \\
\multicolumn{1}{l|}{He et al.\cite{he15}}                 & 40.2 \small{$\pm$0.4}               & 88.6 \small{$\pm$0.7}              & 133.4 \small{$\pm$1.4}              & 206.6 \small{$\pm$1.2}               \\
\multicolumn{1}{l|}{Ours}                 & \textbf{39.2 \small{$\pm$0.4}}               & \textbf{86.4 \small{$\pm$0.6}}            & \textbf{129.4 \small{$\pm$1.0}}              & \textbf{202.2 \small{$\pm$0.5}}             \\
%\multicolumn{1}{l|}{Ours w/ inductive bias}                 & 40.1 \textbf{($\pm$0.1)}             & \textbf{87.6 ($\pm$0.4)}              & 131.6 ($\pm$1.1)             & 204.2 ($\pm$1.1)             \\ 
\hline
\end{tabular}}
\caption{Initialization procedures for best practices model.}
\label{tab:init}
\end{table}

% One should clarify that we refer here to the best possible model, already. May we add the results of init also on the base model, where they should be larger? If it stays so, the inductive bias part should be removed. But there is already a nice story in the paper, so improving the inductive is not a priority. Our priority is the paper and a good story into the ablations.

%%%%%%%%% CONCLUSIONS
\section{Conclusion}

This work has identified, reviewed, and experimentally evaluated best practices for 2-body pose forecasting, to bootstrap research in the mostly unexplored task. Best practices have a large impact on SoA performance, and the novel initialization adds further improvement in performance and stability. Notably, predicting the future of two people in interaction yields better estimates than considering each person separately, so 2-body forecasting is recommended for applications such as sports and collaborative assembly in factories.

%%%%%%%%% REFERENCES
{\small
\bibliographystyle{ieee_fullname}
\bibliography{egbib}
}

\end{document}